\documentclass[letterpaper, 10 pt, conference, twoside]{./iEEEtran}
\usepackage{amsmath,amssymb,amsfonts}
\usepackage{balance}
\usepackage{bm}
\usepackage{lettrine}
\usepackage{enumitem}
\usepackage{hyperref}
\usepackage{color}
\usepackage[table,xcdraw]{xcolor}
\usepackage{tabularx}
\usepackage[caption=false,font=normalsize,labelfont=sf,textfont=sf]{subfig}
\usepackage{booktabs}
\usepackage{graphicx}
\usepackage{multirow}
\usepackage{tikz}
\usepackage{pgfplots}
\usepackage{siunitx}
\usepackage{tabularx}
\usepackage{cite}
\usetikzlibrary{bending, external}

\graphicspath{{./Figures/}}
\DeclareGraphicsExtensions{.pdf,.png,.jpg,.eps,.svg}

\IEEEoverridecommandlockouts
\pgfplotsset{compat=1.16}
\begin{document}
\title{Design and Control of a Coaxial Dual-rotor Reconfigurable Tailsitter UAV Based on Swashplateless Mechanism}
\author{Jinfeng Liang$^{*}$, Haocheng Guo$^{*}$, and Ximin Lyu
\thanks{Manuscript received: August 8, 2025; Revised October 22, 2025; Accepted October 31, 2025.}
\thanks{$^{*}$ \textbf{Jinfeng Liang and Haocheng Guo contributed equally to this work.}}
}

\maketitle
\begin{abstract}
The tailsitter vertical takeoff and landing (VTOL) UAV is widely used due to its lower dead weight, which eliminates the actuators and mechanisms for tilting. However, the tailsitter UAV is susceptible to wind disturbances in multi-rotor mode, as it exposes a large frontal fuselage area. To address this issue, our tailsitter UAV features a reconfigurable wing design, allowing wings to retract in multi-rotor mode and extend in fixed-wing mode. Considering power efficiency, we design a coaxial heterogeneous dual-rotor configuration, which significantly reduces the total power consumption. To reduce structural weight and simplify structural complexity, we employ a swashplateless mechanism with an improved design to control pitch and roll in multi-rotor mode. We optimize the structure of the swashplateless mechanism by adding flapping hinges, which reduces vibration during cyclic acceleration and deceleration. Finally, we perform comprehensive transition flight tests to validate stable flight performance across the entire flight envelope of the tailsitter UAV. For supplementary video see \url{https://youtu.be/Ew_GMr8HCIM}.

\end{abstract}


\section{INTRODUCTION}
\IEEEPARstart{T}{ailsitter} vertical takeoff and landing (VTOL) unmanned aerial vehicles (UAVs) offer distinct advantages due to their ability to transition between multi-rotor mode and fixed-wing mode without additional motors or tilting servos~\cite{bacchini2019electric}. This feature is particularly critical for small-scale VTOLs, as it significantly reduces weight and simplifies manufacturing processes. While exhibiting the aforementioned benefits, small tailsitter UAVs face several key challenges that hinder their development and practical application:
\begin{enumerate}
    \item In multi-rotor mode, the large frontal area substantially makes the vehicle more susceptible to wind disturbance, limiting stability and performance in adverse weather conditions.
    \item Traditional tailsitter designs use propellers with identical diameters. The single-type power unit achieves maximum efficiency solely under a unique working condition, which fails to optimize power efficiency across both multi-rotor and fixed-wing flight modes. This lack of adaptability restricts overall power efficiency.
    \item Coaxial dual-rotor tailsitters with good portability often employ swashplate~\cite{wei2019research,robinson2013design} or thrust-vectoring mechanisms~\cite{gao2024design,garcia2009modeling} for pitch and roll control. However, these solutions introduce phase lag due to limited servo response speeds or small control surfaces~\cite{chen2023swashplateless}. More importantly, from a mechanical standpoint, such torque-producing actuators inherently increase complexity and require extensive maintenance effort.
\end{enumerate}
\begin{figure}[t]
    \centering
    \includegraphics[width=0.9\linewidth]{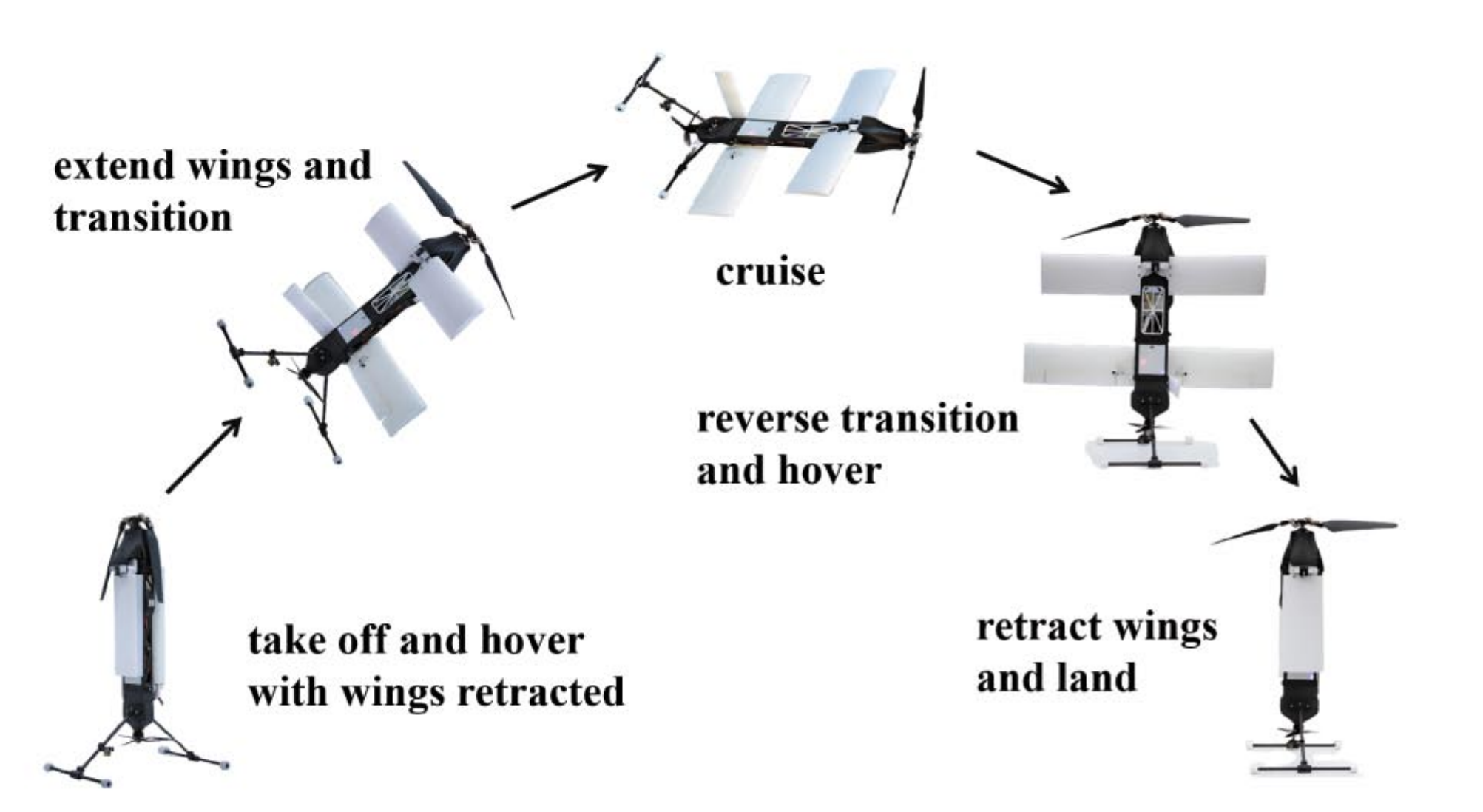}
    \caption{Transition and reconfiguration process of DART: a coaxial dual-rotor reconfigurable tailsitter UAV}
    \vspace{-12pt} 
    \label{fig:wholeproscess}
\end{figure}

To solve issue (1), there are methods based on sensor data fusion~\cite{sun2018wind} and disturbance observer (DOB)~\cite {lyu2018disturbance}. They estimate wind vector (wind speed and wind direction) and mitigate tracking errors through disturbance compensation. However, such software-based approaches are prone to inducing actuator saturation. We propose a hardware-based reconfigurable mechanism to address this issue. The aircraft's frontal area is reduced by $66.2\%$ through retracting all wings to address wind disturbance. 

To address issue (2), common power optimization methods include variable-pitch propeller (VPP) configuration~\cite{simmons2022efficient} and heterogeneous multi-propeller configuration~\cite{wang2015modeling}. VPP maintains power consumption at a low level across various speed conditions by adjusting the collective pitch to the relative wind vector. However, VPP requires an additional mechanical structure that is complex and costly to maintain,  which is expensive for a small tailsitter UAV. Heterogeneous multi-set propeller configurations present a viable approach to optimizing power efficiency without the need for a VPP mechanism. Thus, we adopt a coaxial heterogeneous dual-rotor (CHD) configuration, featuring a fore-rotor with a large-diameter motor directly driving a swashplateless propeller and an aft-rotor with a small-diameter motor coupled to a fixed-pitch propeller. Specifically, different propellers are utilized in distinct flight modes: both the large-diameter and small-diameter propellers are employed in multi-rotor mode, while only the small-diameter propeller is active in fixed-wing mode.

To solve issue (3), we employ the swashplateless mechanism (SPLM) as the roll and pitch actuator. We optimize the SPLM structure, resulting in reduced vibration as verified by bench tests. Utilizing an efficient controller framework, it becomes possible to utilize only two actuators to govern six degrees of freedom in multi-rotor mode.

To validate the proposed design, we developed a prototype named the \textbf{D}u\textbf{a}l-rotor \textbf{R}econfigurable \textbf{T}ailsitter (DART). As shown in Fig.~\ref{fig:wholeproscess}, the vehicle retracts wings during hovering, and extends wings for transition and cruise flight.
A series of flight experiments and simulation tests were conducted to demonstrate the feasibility and effectiveness of the proposed configuration. 
\begin{figure}[t]
\vspace{12pt}
	\centering
\includegraphics[width=0.9\columnwidth]{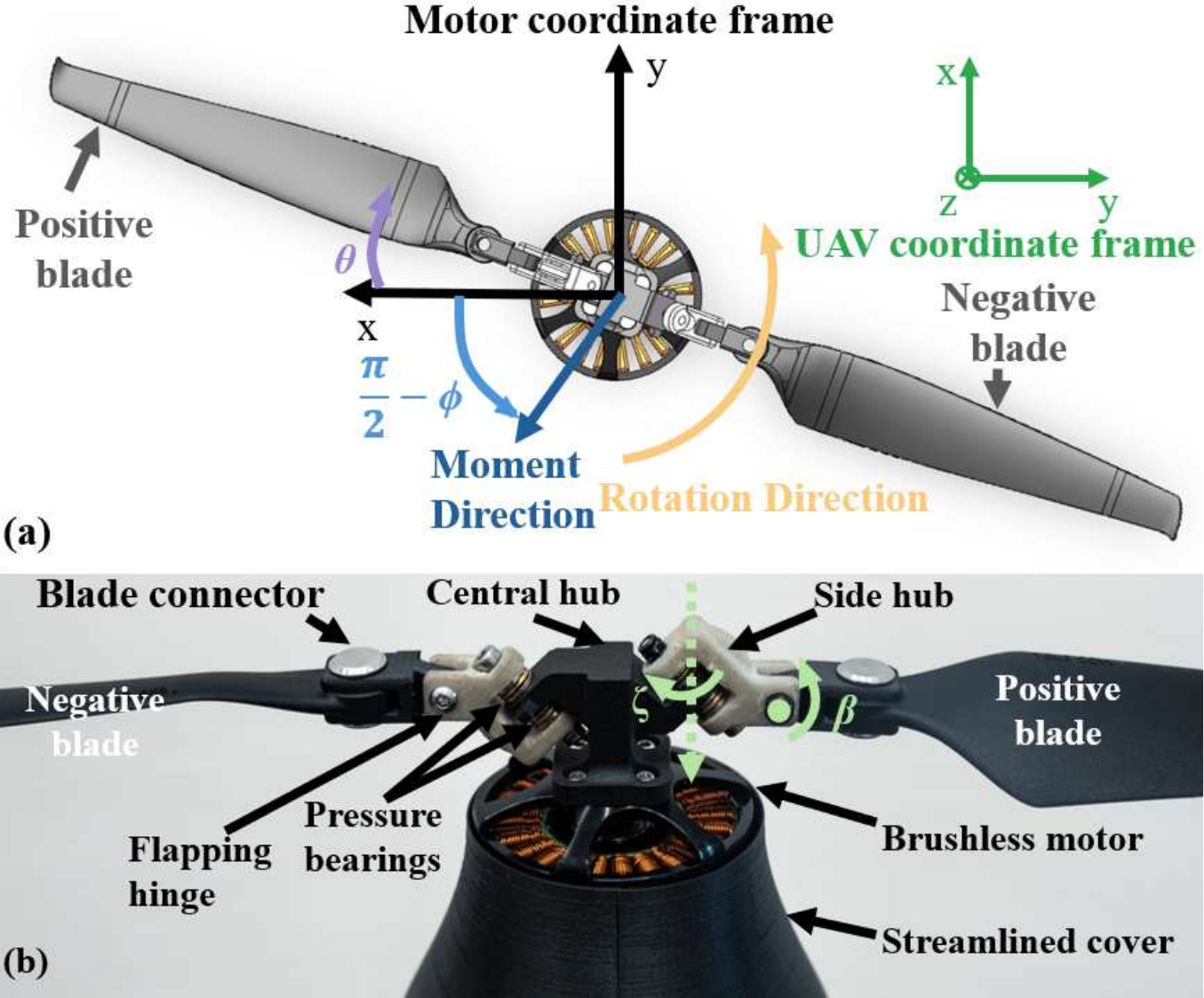}
 \caption{(a) Definition of coordinate Frames for SPLM and UAV (Top View). (b) The components of optimized SPLM.
		\label{fig:slm}}
	\vspace{-20pt}
\end{figure}
To be specific, our research makes the following key contributions:
\begin{enumerate}
\item \textbf{Reconfigurable Tailsitter}: To mitigate wind disturbances, we developed a reconfigurable mechanism for wing retraction in multi-rotor mode, which reduced the vehicle's susceptibility under a 5 m/s wind. During the transition, the DART extends its wings and speeds up to reach sufficient airspeed before switching to fixed-wing mode.
\item \textbf{CHD configuration}: To achieve enhanced power efficiency, we propose a CHD configuration, comprising a fore-rotor with large-radius blades and an aft-rotor with small-radius blades. Based on empirical aerodynamic formulas for propeller thrust and power in varying advance ratios, the implementation of a CHD configuration can reduce average power consumption by $29.2\%$ versus a homogeneous configuration when the hover time accounts for $20\%$ of the total mission time, a typical VTOL mission profile.
\item \textbf{Optimized SPLM}: To reduce the vibration of the SPLM, we propose a novel SPLM design featuring a couple of flapping hinges. The design not only avoids the direct alternating stress transmission, but also eliminates nonlinear terms in the dynamic function. Experiments demonstrate that the torque sampling data of the optimized SPLM exhibited a 62.9$\%$ reduction compared to that without flapping hinges, and further reveal an average vibration energy reduction of 6 dB/Hz.
\item \textbf{Efficient control system for underactuated SPLM VTOL}:
In multi-rotor mode, we use only two actuators to control 6 DOF motion with less vibration. With the high-bandwidth characteristics of the SPLM, our proposed control framework enables stable control of attitude and position.
\end{enumerate}
\vspace{-4pt}
\section{RELATED WORK}
\subsection{Reconfigurable Wings Tailsitter}
Vourtsis et al.~\cite{vourtsis2021robotic} proposed an insect-inspired compact aircraft with elytra-equipped folding wings, which generates additional lift through a biplane configuration to offset weight. Kevin et al.~\cite{ang2015design} designed a tailsitter with reconfigurable wings named U-lion, which utilizes a single servo-driven four-bar linkage design to achieve wing retracting and extending. However, their larger frontal area still poses challenges for hover control in windy environments. Cai et al.~\cite{cai2024development} proposed a bird-wing-inspired wing folding mechanism that uses a linear actuator. This mechanism enables the wings to retract on both sides by rotating around specific axes. Although this configuration substantially reduces frontal area, they adopt a conventional aerodynamic layout. This means their long wingspan results in inadequate stowage and poor space utilization when retracted. In addition, a horizontal stabilizer that generates negative lift will reduce aerodynamic efficiency. We adopted a more aerodynamically efficient tandem wing layout for which we designed a servo-based folding mechanism. This enables a compact form factor for portability and inherent stability against wind disturbances in multi-rotor mode.

\vspace{-4pt}
\subsection{Propulsion Efficiency Optimization}
The utilization of variable-pitch mechanisms to adaptively adjust the collective pitch of propellers based on the relative wind vector is a prevalent method for enhancing the efficiency of propellers.  Duan et al.~\cite{duan2024optimization} proposed a control framework for variable-pitch propellers, which can minimize energy consumption and sustain aircraft stability simultaneously. However, variable-pitch propeller systems inherently require additional actuators and associated intricate mechanical linkages, leading to an additional weight and high-maintenance. Henderson et al.~\cite{Henderson2020optcontrol} proposed an adaptive control algorithm to minimize electrical consumption for variable-pitch mechanisms, but high computational costs and actuator precision needs make it uneconomical in practical engineering.
A dual-system VTOL~\cite{gu2017development} is widely used because of its ease of deployment. It uses two independent propulsion systems for multi-rotor mode and fixed-wing mode. By optimizing the propulsion system for the two flight modes separately, the average efficiency of VTOL aircraft in both multi-rotor mode and fixed-wing mode can be improved. Inspired by this, our tailsitter utilizes two sizes of propellers, each optimized for hover and level flight conditions, respectively.

\subsection{SPLM}
The SPLM is a passive cyclic variable pitch mechanism which uses asymmetric lag-pitch hinge design~\cite{paulos2018scalability}. Sinusoidal modulation induces periodic adjustments in blade pitch to generate a specific directional torque. There are flap-hinge-less versions and flap-hinge versions of SPLM. Paulos and Yim et al.~\cite{paulos2013underactuated,paulos2015flight} were the first to design a flap-hinge-less version and applied it to a coaxial dual-rotor configuration. Despite using two blades larger than 30 cm, the aircraft's takeoff weight is limited to only 358 g. Chen et al.~\cite{Chen2023ASS} optimized the mechanism, applying a single SPLM to a 1.23 kg autonomous drone and validating its robustness and tracking accuracy under various environments. However, the rigid connection between the hub and the blade in the flap-hinge-less configuration will transmit alternating stress, resulting in harmonic oscillation. Paulos was the first to integrate flapping hinges into the SPLM, with hinges cross-connected to the lag-pitch hinge~\cite{paulos2018emulating}. This configuration enabled thrust vectoring and held potential for vibration improvement. However, it results in a coupling between the two motions. The stiffness matrix in the system's state-space equation becomes unsteady with nonlinear terms. This asymmetry causes the system response phase difference between the two blades, leading to instability of the system response. As shown in Fig.~\ref{fig:slm}, based on their work, we move the flapping hinge to the junction of the blade and the side hub to decouple the flapping and lag-pitch motions kinematically.

\section{HARDWARE DESIGN}
\label{sec:hardware_design}
\subsection{UAV Configuration}
\begin{figure}[t]
    \centering
        \vspace{3pt}
    \includegraphics[width=0.9\linewidth]
    {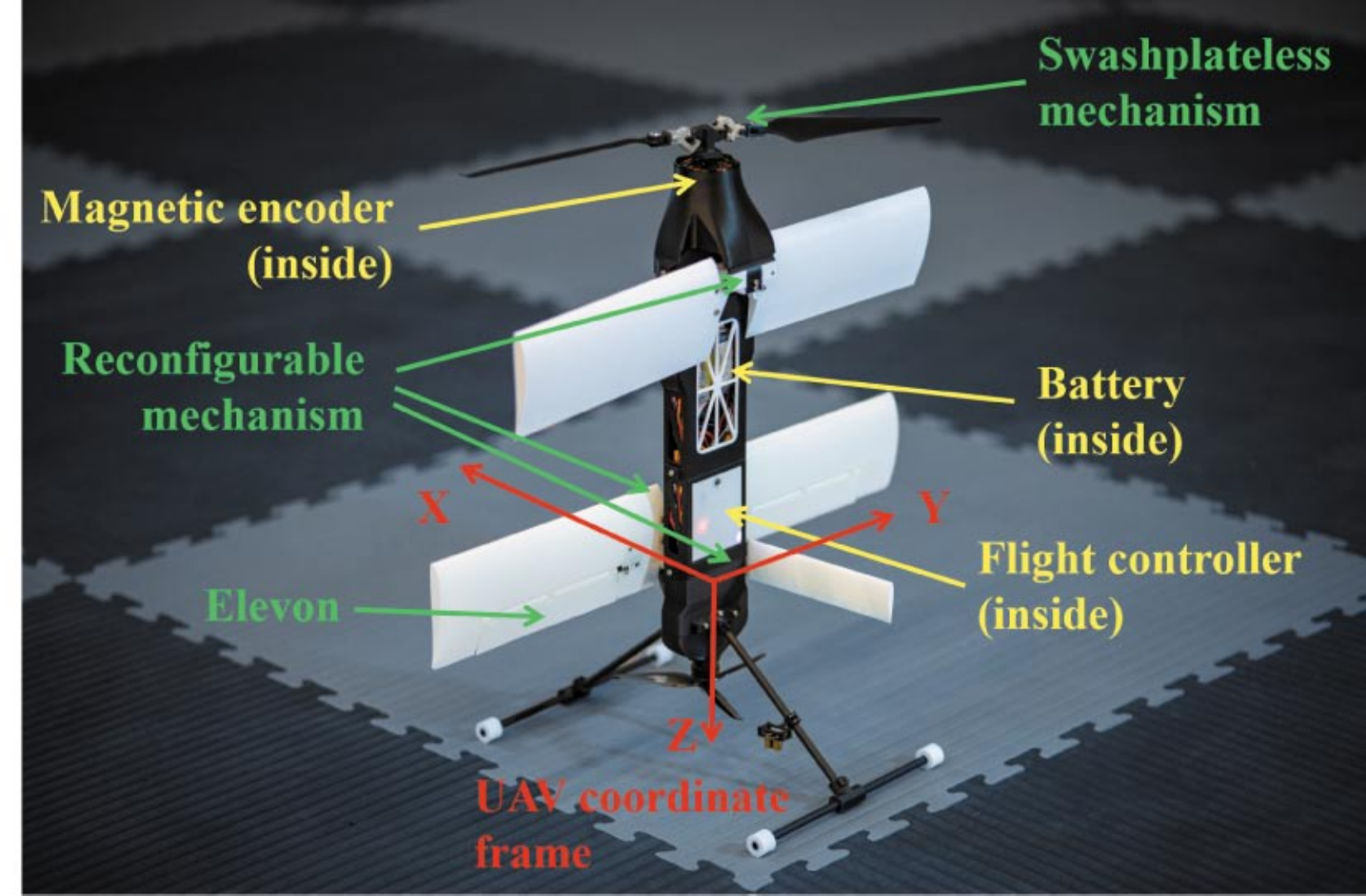}
    \caption{Layout of the DART}
    \label{fig:system_intro}
    \vspace{-18pt}
\end{figure}

The DART, shown in Fig.~\ref{fig:system_intro}, was designed for reduced susceptibility to wind disturbances in multi-rotor mode and high cruise efficiency in fixed-wing mode. To enhance hover stability in windy conditions, DART retracts its wings during multi-rotor mode. The folding mechanism aligns wings with the fuselage outline, effectively reducing lateral wind exposure area by $66.2\%$. This geometric adaptation significantly lowers crosswind-induced disturbance.
To achieve higher efficiency in the whole flight process, DART employs a CHD configuration. The fore-rotor 16-inch propellers maintain efficiency during low-speed cruise, while the aft-rotor 7-inch propellers maintain efficiency during high-speed flight through their reduced flow separation characteristics. This dual-propeller arrangement decreases the thrust decay across operational speed ranges.
To balance flight performance with structural compactness, a tandem wing configuration was selected because it allows all wings to generate positive lift and use high-lift airfoils without trim constraints. As a result, the DART achieves a higher lift coefficient compared to the conventional layout and flying wing layout.
The tandem wing layout also offers flexibility in wing placement. The front wing is positioned above the fuselage, while the rear wing is below it. The left and right wings are staggered to prevent interference during folding. Elevons are installed on the rear wing to control roll and pitch in transition mode and fixed-wing mode.
\subsection{Aerodynamic Design}
\begin{figure}[h]
    \centering
    \vspace{-12pt}
    \includegraphics[width=0.85\linewidth]{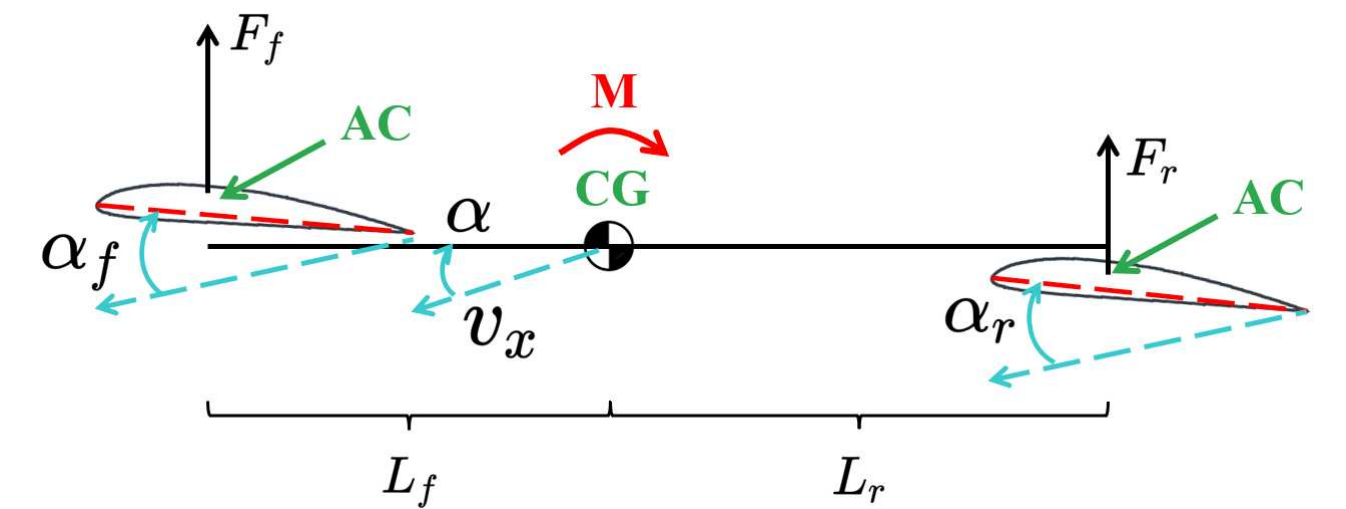}
    \vspace{-6pt}
    \caption{Lever model for tandem wings aircraft}
    \vspace{-6pt}
    \label{fig:lever_model}
\end{figure}

Airfoil selection influences trim characteristics and aerodynamic performance for tandem wing configurations. A lever analogy model is established for tandem wings, as depicted in Fig.~\ref{fig:lever_model},
where the aerodynamic center (AC) of the front and rear wings act as the ends of a lever, with the aircraft's center of gravity (CG) as the fulcrum. $\alpha$ is the angle of attack (AoA) of the plane.

The lift calculation formula is given by 
\begin{equation}
\label{equ:airforece}
F =\frac{1}{2}\rho Sv^2Cl,
\end{equation}
where $\rho$ denotes air density, $S$ is wing area, $v$ is freestream velocity, and $Cl$ is the lift coefficient. $Cl$ normally demonstrates linear characteristics between -5° and 10° AoA, expressed as
\begin{equation}
\label{equ:torque}
Cl = k\alpha_{airfoil}+Cl_{0},
\end{equation}
where $k$ denotes the lift curve slope, $\alpha_{airfoil}$ denotes the AoA of the airfoil.$C_{\text{$l$0}}$ represents the lift coefficient at zero AoA.

During flight, the forces acting on the UAV can be expressed as:
\begin{equation}
    \left\{
    \label{equ:stability_1}
    \begin{aligned}
    F_{f}&=\frac{1}{2}\rho S_fv^2(k_f\alpha_{f}+Cl_{f0}),\\
    F_{r}&=\frac{1}{2}\rho S_rv^2(k_r\alpha_{r}+Cl_{r0}),\\
    M&=F_{r} L_r-F_{f} L_f,
    \end{aligned}
    \right.
\end{equation}
where $F_{f}$ and $F_{r}$ are the lift of the front wing and rear wing, respectively. $M$ is the resultant moment about CG. Subscripts $f$ and $r$ denote parameters for the front wing and rear wing, respectively: $S_f$, $S_r$ (wing areas), $k_f$, $k_r$ (lift curve slopes), $\alpha_{f}$, $\alpha_{r}$ (AoA of the airfoil), $L_f$ and $L_r$ (distances between the airfoil's AC to the vehicle's CG).  $Cl_{f0}$ and $Cl_{r0}$ are the lift coefficients of the front wing and the rear wing at zero AoA of the airfoil. 

The AoA of the front wing and rear wing are decomposed as \eqref{equ:stability_5},
\begin{equation}
\left\{
\label{equ:stability_5}
    \begin{aligned}
\alpha_f = \alpha_{f0} + \Delta\alpha,\\
\alpha_r = \alpha_{r0} + \Delta\alpha,
\end{aligned}
    \right.
\end{equation}
where $\alpha_{f0}$ and $\alpha_{r0}$ denote the respective equilibrium values during cruise flight, at which the aircraft maintains torque balance without control input. $\Delta\alpha$ represents the instantaneous deviation of the AoA from the cruise condition.

In our design, both the front wing and the rear wing adopt the same airfoil, which means:
\begin{equation}
\left\{ 
\label{equ:stability_2}
\begin{aligned}
k_f&=k_r,\\
Cl_{f0}&=Cl_{r0}.
\end{aligned}
\right.
\end{equation}

During cruise, the UAV must maintain equilibrium. A resultant moment about the CG is required to restore this state after a disturbance. Especially, the resultant moment should strictly monotonically decrease with $\Delta\alpha$, which means:
\begin{equation}
\left\{ 
\begin{aligned}
&M = 0, &&  \Delta \alpha = 0, \\
&M \cdot \Delta \alpha < 0, && \Delta \alpha \neq 0,\\
&\dfrac{\partial M}{\partial (\Delta \alpha)}<0,&& \alpha_f \in [-5^\circ, 10^\circ], \ \alpha_r \in [-5^\circ, 10^\circ].
\end{aligned}
\right.
\label{equ:stability_3}
\end{equation}

Assuming the lift coefficient of the rear wing is not affected by the downwash of the front wing, substitute \eqref{equ:stability_5}, \eqref{equ:stability_2}, and \eqref{equ:stability_3} into \eqref{equ:stability_1} to derive the longitudinal static stability condition as follows:
\begin{equation}
\left\{ 
\label{equ:stability_4}
\begin{aligned}
\alpha_{r0}&<\alpha_{f0},\\
S_fL_f&<S_rL_r.
\end{aligned}
\right.
\end{equation}

\begin{figure}[h]
    \centering
    \vspace{-6pt}
    \includegraphics[width=1\linewidth]{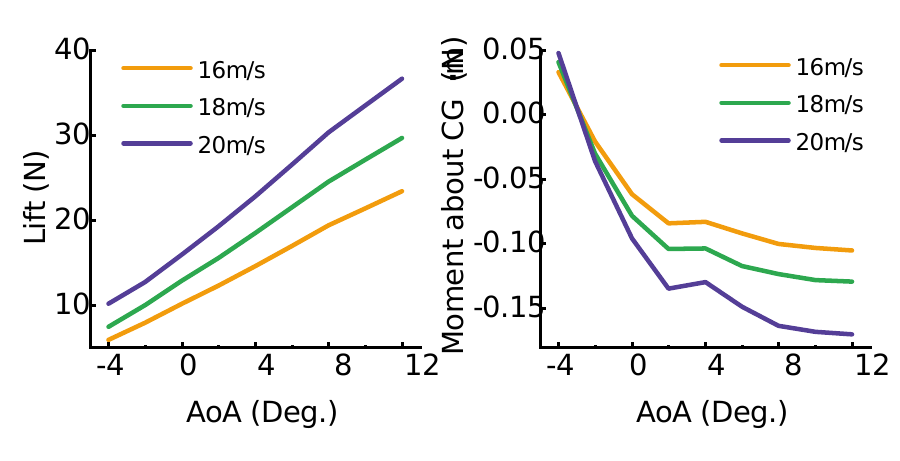}
    \vspace{-20pt}
    \caption{Curve of lift and moment versus AoA}
    \vspace{-6pt} 
    \label{fig:lift-drag-ratio}
    	\vspace{-0pt}
\end{figure}

Parametric CAD modeling and CFD analysis enabled iterative optimization of DART's geometry (Table~\ref{tab:wing_parameters}).
\begin{table}[b]
\centering
\vspace{-6pt}
\caption{Parameters of Front Wing and Rear Wing}
\begin{tabular}{cc}
\toprule
\textbf{Name} & \textbf{Parameter} \\ \midrule
Airfoil of the front wing and rear wing & MH114\\
Wing span of the front wing & 480 mm \\
Chord length of the front wing & 100 mm \\
Incidence angle of the front wing & 4.5° \\
Wing span of the rear wing & 560 mm \\
Chord length of the rear wing & 100 mm \\
Incidence angle of the rear wing & 2° \\
\bottomrule
\end{tabular}
\label{tab:wing_parameters}
\end{table}
Aerodynamic data (Fig.~\ref{fig:lift-drag-ratio}) 
show that the DART possesses adequate aerodynamic performance and achieves stable aerodynamic characteristics. From -4° to 12° AoA, the lift increases strictly with AoA. The restoring moment generally increases with rising AoA, but shows a distinct decrease between 2°and 4° AoA. This particular behavior stems from front-rear wing aerodynamic coupling.
\vspace{-6pt}
\subsection{Optimized SPLM}
As shown in Fig.~\ref{fig:slm}, our optimized SPLM is based on PULSAR~\cite{Chen2023ASS}, which comprises a central hub with asymmetric hinges and a side hub directly connecting the blade. A pair of pressure bearings is added to each lag-pitch hinge to further reduce friction. A flapping hinge was introduced to connect the side hub and the blade. Unlike previous designs~\cite{Gemini2,paulos2018emulating}, our design decouples the blade flapping motion and lag-pitch motion. As presented in~\cite{paulos2018scalability}, the dynamical model for the rotor system can be simplified as: 
\begin{equation}
    \mathbf{M}\mathbf{x}'' + \mathbf{C}\mathbf{x}' + \mathbf{K}\mathbf{x} = \begin{bmatrix} \dfrac{u}{a \cdot \sigma} \\ 0 \\ 0 \end{bmatrix}.
\end{equation}
The model parameters are defined as follows: $\mathbf{M} \in \mathbb{R}^{3\times3}$ is the constant inertia matrix determined by SPLM geometry and mass distribution; $\mathbf{C} \in \mathbb{R}^{3\times3}$ represents the aerodynamic damping matrix dependent on blade parameters and average motor speed; $\mathbf{K} = \mathbf{K}_c + \mathbf{K}_\beta \in \mathbb{R}^{3\times3}$ denotes the stiffness matrix comprising constant and nonlinear components. The state vector is given by $\mathbf{x} = [\theta, \zeta, \beta]^T$, as shown in Fig.~\ref{fig:slm}, $\theta$ is the motor angle (rad, positive blade aligns with positive $x$-axis in motor coordinate frame represents 0 rad) sampled by the encoder, $\zeta$ is the lag angle (rad), and $\beta$ is the flapping angle (rad). The system inputs include $u$ for motor modulation input and $a$ for blade section lift curve slope. Additionally, $\sigma$ represents the rotor solidity (dimensionless), defined as $\sigma = Nc/\pi R$, where $N$, $c$, and $R$ are the blade count, chord length, and rotor radius, respectively.

 Both the $\mathbf{M}$ matrix and $\mathbf{C}$ matrix are constant matrices, while for SPLM, the $\mathbf{K}$ matrix consists of a constant part $\mathbf{K}_c$ and a state-dependent part $\mathbf{K}_\beta$:
\begin{equation}
    \mathbf{K}_\beta = \frac{1}{8}
    \begin{bmatrix}
        \Phi_{3/4} \\
        -\Phi_{3/4}\left(1 - \dfrac{4e}{3}\right) \\
        \dfrac{4e}{3} - 1
    \end{bmatrix}
    \begin{bmatrix}
        0 & \dfrac{\Delta\alpha_b}{\Delta\zeta} & 0
    \end{bmatrix},
\end{equation}
the parameters are defined as follows: $ \Phi_{3/4}$ denotes the downwash angle at the 3/4 spanwise station (rad),  $\alpha_b$ represents the blade pitch (rad), and $e$ denotes the flapping hinge location relative to the blade tip radius. \({\Delta\alpha_b}/{\Delta\zeta}\) is expressed by:
\begin{equation}
    \dfrac{\Delta\alpha_b}{\Delta\zeta} = \tan\left(\beta + \dfrac{\pi}{4}\right).
\end{equation}
In our design, the kinematic decoupling of the flapping and lag-pitch motions results in the term \({\Delta\alpha_b}/{\Delta\zeta}\) being equal to 1. Thus, \( \mathbf{K}_{\beta} \) reduces to a constant matrix. This design feature is expected to linearize the originally existing nonlinear term in the system, offering theoretical guidance for enhancing the stability of the system's dynamic response.

\subsection{Reconfigurable Wing Design}
\begin{figure}[h]
	\vspace{-6pt}
	\centering
	\includegraphics[width=0.9\columnwidth]{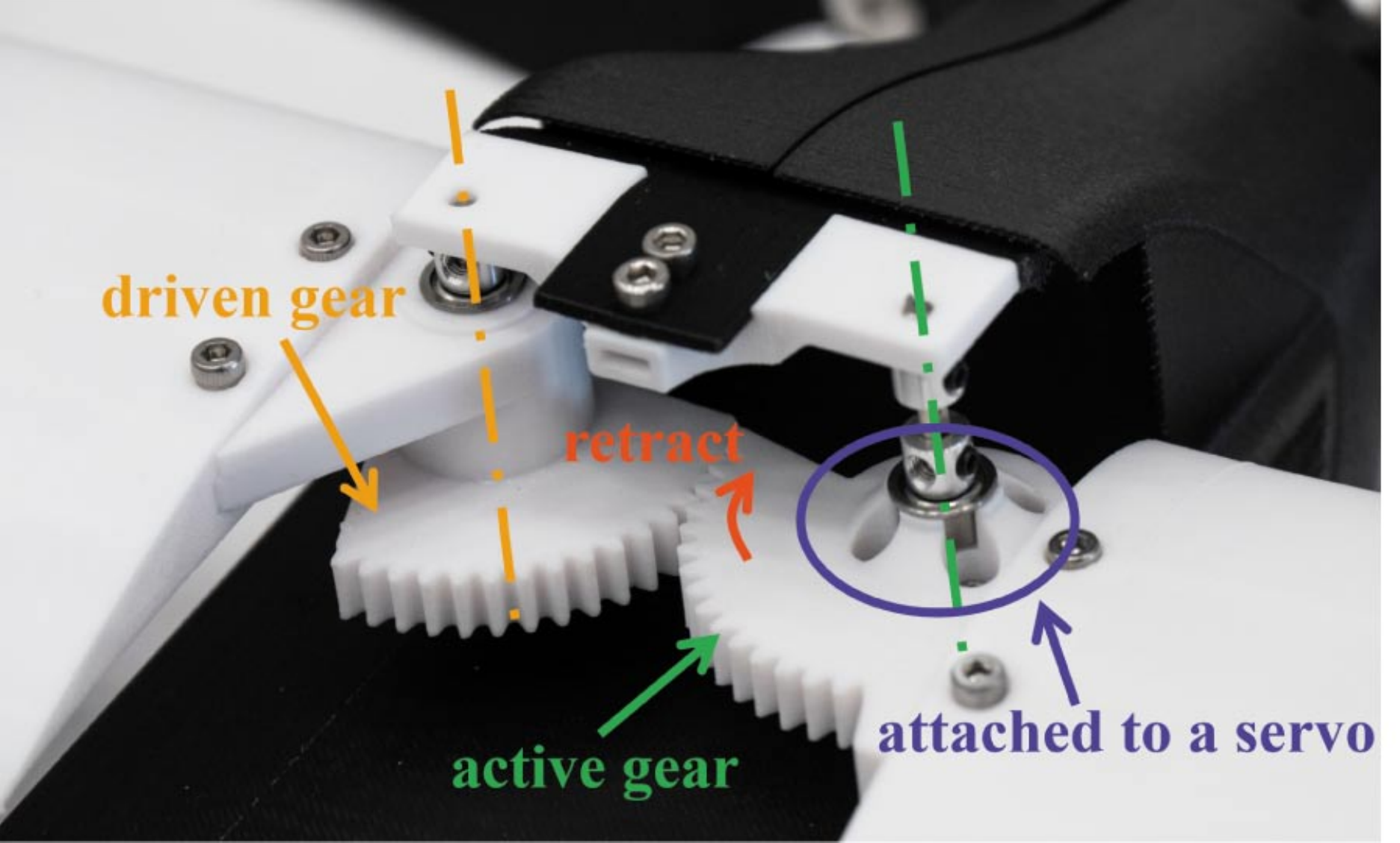}
 \caption{Gear mechanism for reconfigurable wing design
		\label{fig:gear}}
            \vspace{-6pt} 
\end{figure}

We proposed a reconfigurable wing actuation system using a gear mechanism for both front wings and rear wings. As shown in Fig.~\ref{fig:gear}, the gear system is based on a standard involute straight gear, with a module of 1 and 50 teeth. An active gear is fixedly attached to a servo and actuates the driven gear. The vertical tail is also directly driven by a servo.
\vspace{3pt}
\section{CONTROL}
\subsection{Control System Overview}
\begin{figure*}[h]
\vspace{3pt}
    \centering
    \includegraphics[width=1.93\columnwidth]{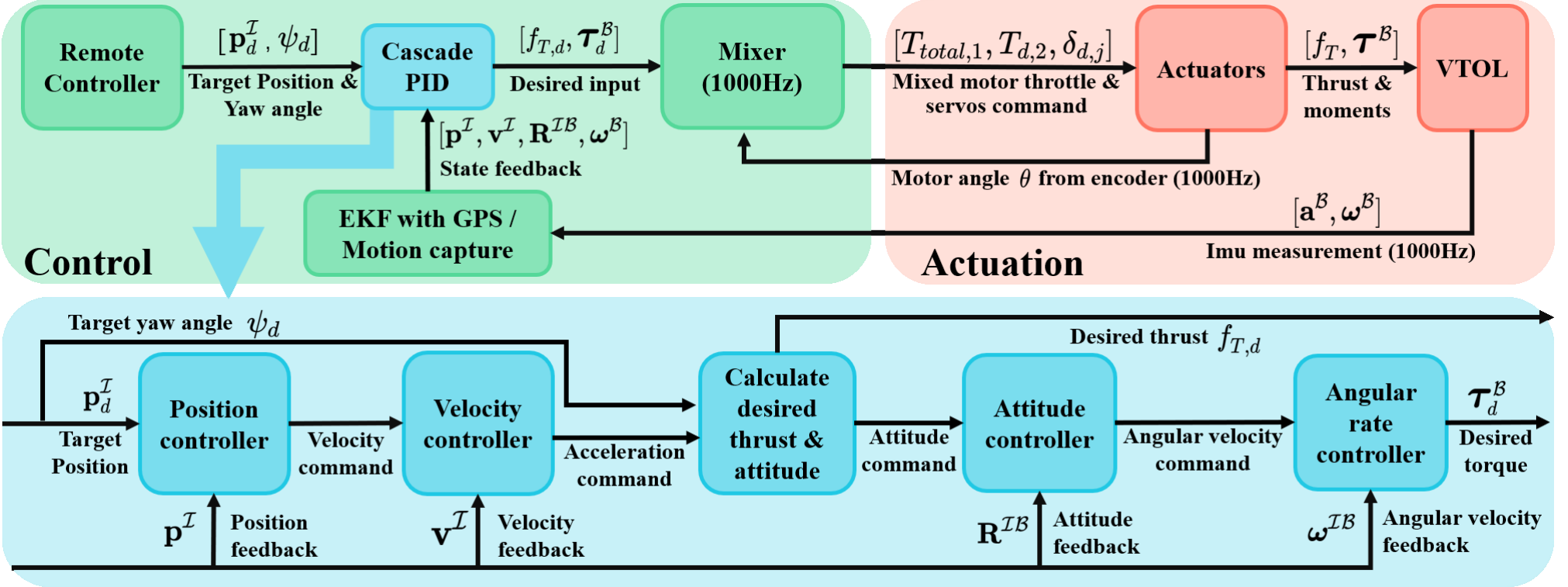}
    \caption{Control system overview}
    \label{fig:example}
        \vspace{-12pt} 
\end{figure*}

The flight control system is illustrated in Fig.~\ref{fig:example}. The input commands are position and yaw setpoints generated by a remote controller.
\label{sec:modeling_and_control}
Control system adopts the standard cascaded control architecture based on PX4, with proportional (P) controllers for position and attitude loops, and proportional-integral-derivative (PID) controllers for velocity and angular rate loops. Specifically, the tracking frequencies for yaw rate and roll pitch rate are differentially allocated at 200 Hz and 1000 Hz, which intends to rationally utilize the mid-frequency range of the SPLM's wide bandwidth. A customized mixer module, specifically designed for the SPLM, operates at 1000 Hz to synchronize with the sampling frequency of magnetic encoders. The mixer generates the throttle command for fore-rotor and aft-rotor, as well as the servo commands for the elevons. The servo commands are only active in fixed-wing mode. The sensing module integrates external positioning sensors, such as a motion capture system or GPS, along with an onboard IMU embedded in the flight controller. The measurements are fused through the EKF module to generate estimated states, including position, velocity, attitude, and angular rate.
\subsection{Mixer design}
The mixer equations are formulated as:
\begin{equation}
\label{equ:mixer1}
\setlength{\arraycolsep}{1pt}
 \begin{bmatrix}
             f_{T,d} \\
            \tau_x^B \\
            \tau_y^B \\
            \tau_z^B
        \end{bmatrix}
        =
        \setlength{\arraycolsep}{6pt}
        \begin{bmatrix}
            1        & 1    &     0 & 0 & 0 & 0 & 0\\
            0        & 0    & 1 & 0 & 0 & 0 & 0\\
            0        & 0 & 0 & 1 & 1 & 0 & 0\\
            0 & 0 & 0 & 0 & 0 & 1 & 1\\
        \end{bmatrix}
        \begin{bmatrix}
            f_{T,1} \\
            f_{T,2} \\
            M_{s,x} \\
            M_{s,y} \\
            M_{\delta,y} \\
            M_{\delta,z}\\
            M_{T,z}
        \end{bmatrix},
\end{equation}
where $f_{T,d}$ denotes the desired thrust, $\{{\tau}_x^B,{\tau}_y^B,{\tau}_z^B\}$ is the desired moment. The thrust of motor $i$ ($i = 1, 2$) is represented by  $f_{T,i}$. $M_{s,x}$ and $M_{s,y}$ are the components of desired SPLM moment vector $\mathbf{M}_s$ along the body's $x$-axis and $y$-axis, respectively.  $M_{\delta,y}$ and $M_{\delta,z}$ are the components of  desired elevon moment vector $\mathbf{M}_{\delta}$  along the body's $y$-axis and $z$-axis. $M_{T,z} $ represents the yaw moment generated by the differential torque of two motors. We assume a linear mapping from the desired control input to the actual actuator input.  The mapping function is as follows:
\begin{equation}
\label{equ:mixer2}
 \begin{bmatrix}
            f_{T,1} \\
            f_{T,2} \\
            M_{s,x} \\
            M_{s,y} \\
            M_{\delta,y} \\
            M_{\delta,z}\\
            M_{T,z}
        \end{bmatrix}
        =
         \setlength{\arraycolsep}{0.5pt}
        \begin{bmatrix}
            C_{T,1} & 0 & 0 & 0 & 0 & 0 \\
            0 &  C_{T,2} & 0 & 0 & 0 & 0 \\
            0 & 0 & C_m & 0  & 0 & 0\\
            0 & 0 & 0 & C_m & 0 & 0 \\
            0 & 0 &0 & 0 & k_{e,y} & k_{e,y} \\
            0 & 0 &0 & 0 & k_{e,z} & -k_{e,z}\\
            -K_{t,1} & K_{t,2} &0 & 0 &0 & 0
        \end{bmatrix}
        \begin{bmatrix}
            T_{d,1} \\
            T_{d,2} \\
            M_{d,x} \\
            M_{d,y} \\
            \delta_{d,1} \\
            \delta_{d,2}
        \end{bmatrix}.
\end{equation}
The throttle of motor $i$ ($i = 1, 2$) is represented by $T_{d,i}$. $M_{d,x}$ and $M_{d,y}$ are the components of desired SPLM acceleration and deceleration signal $\mathbf{M}_d$ along the body's $x$-axis and $y$-axis. The command sent to elevon servo $j$ ($j = 1, 2$) is denoted by $\delta_{d,j}$. The proportion coefficient $C_m$ is the conversion from the amplitude of the SPLM sinusoidal modulation signal to the actual moment. $C_m$ is calibrated by bench test. $k_{e,y}$ and $k_{e,z}$ are proportion coefficients that describe the conversion from servo command to elevon moment along the body's $y$-axis and $z$-axis. $C_{T,i}$, $K_{t,i}$ are the thrust coefficient and moment coefficient of motor $i$, which can be calibrated by bench test.

After combining \eqref{equ:mixer1} and \eqref{equ:mixer2}, the equation is underdetermined. Considering the actual flight situation, moment distribution is governed by $\lambda \in [0,1]$, where $\lambda$ is the rotor's contribution ratio to the total moment, and $(1-\lambda)$ is allocated to the elevons. Thus, applying $\lambda$ to the underdetermined equation and left-multiplying with the inverse coefficient matrix, the desired actuator inputs are determined as follows:
\begin{equation}
\label{equ:mixer}
 \begin{bmatrix}
            T_{d,1} \\
            T_{d,2} \\
            M_{d,x} \\
            M_{d,y} \\
            \delta_{d,1} \\
            \delta_{d,2}
        \end{bmatrix}
        =
        \begin{bmatrix}
            \eta & 0 & 0 & \gamma \\
            \kappa & 0 & 0 & \delta \\
            0 & \frac{1}{C_m} & 0 & 0 \\
            0 & 0 & \frac{\lambda}{C_m} & 0 \\
            0 & 0 & \frac{1-{\lambda}}{2k_{e,y}} & \frac{1-{\lambda}}{2k_{e,z}} \\
            0 & 0 & \frac{1-{\lambda}}{2k_{e,y}} & \frac{{\lambda}-1}{2k_{e,z}}
        \end{bmatrix}
        \begin{bmatrix}
            f_{T,d} \\
            \tau_x^B \\
            \tau_y^B \\
            \tau_z^B
        \end{bmatrix}.
\end{equation}
The parameters are defined as follows:
\begin{align}
\eta &\triangleq \frac{K_{t,2}}{C_{T,1}K_{t,2}+C_{T,2}K_{t,1}}, \label{eq:alpha} \\
\kappa &\triangleq \frac{K_{t,1}}{C_{T,2}K_{t,1}+C_{T,1}K_{t,2}}, \label{eq:beta} \\
\gamma &\triangleq \frac{-{\lambda}C_{T,2}}{C_{T,1}K_{t,2}+C_{T,2}K_{t,1}}, \label{eq:gamma} \\
\delta &\triangleq \frac{{\lambda}C_{T,1}}{C_{T,2}K_{t,1}+C_{T,1}K_{t,2}}. \label{eq:delta}
\end{align}

$T_{d,2}$, as the output of the aft-rotor, $\delta_{d,1}$ and $\delta_{d,2}$, as the outputs of the two elevons respectively, are all sent to the low-level controller for execution. However, $T_{d,1}$ and $\mathbf{M}_d$ are first processed by a customized mapping layer before being sent to the fore-rotor. Mapping layer follows the SPLM principle, generating moment through single-motor modulation for passively hinged blades. A steady throttle ($T_{d,1}$) generates thrust, while a sinusoidal throttle (phase-locked to rotor angle) is superimposed, making blades cyclically change pitch: As shown in  Fig.~\ref{fig:slm}(b), for example, positive blade increases pitch when bending backward, negative one decreases. This creates uneven lift, generating roll and pitch moments.  Sine amplitude adjusts moment magnitude, phase adjusts direction. The corresponding mathematical equation can be written as:
\vspace{-5pt}
\begin{equation}
\label{equ:Ttotal}
T_{total,1} = T_{d,1} +||\mathbf{M}_d|| \sin\left(\theta + \phi - \beta_{\text{delay}}\right).
\end{equation}
\vspace{-2pt}
$T_{total,1}$ denotes the fore-motor command transmitted to the ESC. $||\mathbf{M}_d||$ is the norm of $\mathbf{M}_d$. As shown in  Fig.~\ref{fig:slm}, \(\phi\) represents the modulation phase calculated based on the direction of $\mathbf{M}_d$.  \(\beta_{\text{delay}}\) is the phase lag offset caused by blade inertia, which varies with the change of the motor's average rotational speed \(\omega\). Through bench testing, \(\beta_{\text{delay}}\) is calibrated across motor speeds.

\section{EXPERIMENTS AND RESULTS}
\label{sec:Experiment}
\subsection{Wind Disturbance Rejection Verification}
\label{subsec:wind_resis_ver}
\begin{figure}[h]
    \vspace*{-6pt} 
    \centering
    \includegraphics[width=0.95\linewidth]{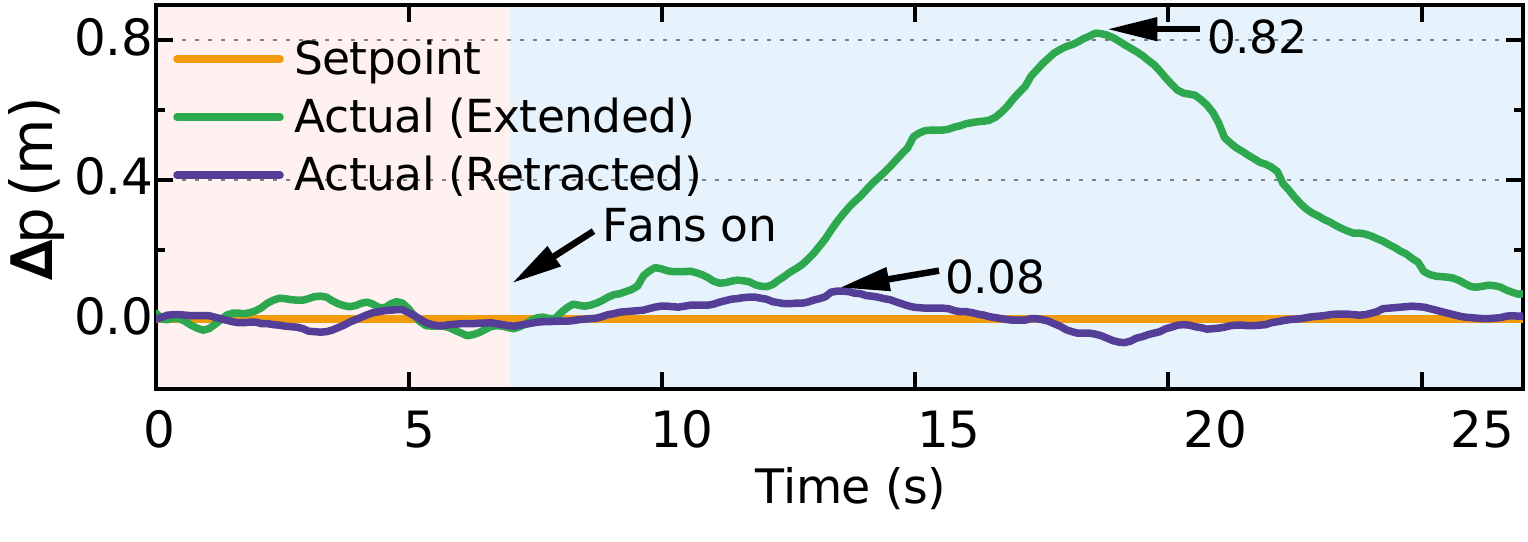}
       \vspace{-6pt} 
    \caption{Position tracking errors of DART with wings extended and retracted under steady wind disturbance}
    \label{fig:wind_resistance}
    \vspace{-18pt}
\end{figure}

This section aims to evaluate the position tracking performance of the DART with retracted wings configuration versus extended wings configuration under gusty wind during multi-rotor mode.
\begin{figure}[h]
    \centering
    \vspace{-12pt} 
    \includegraphics[width=0.9\linewidth]{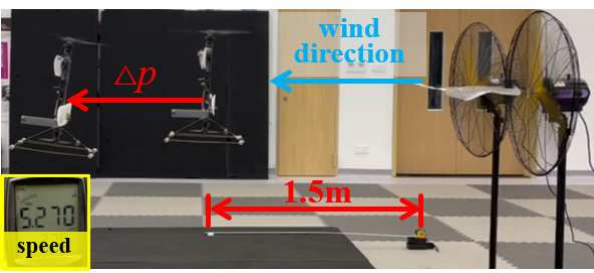}
    \caption{Wind disturbance rejection capability comparison experiment setup of DART}
    \vspace{-8pt}
    \label{fig:wind_setup}
\end{figure}
As illustrated in Fig.~\ref{fig:wind_setup}, the DART hovers 1.5 meters in front of the fans after take-off. Once the vehicle is stabilized in hover, the fans are energized, causing the aircraft to be displaced by the wind. Position tracking error $\Delta p$ is obtained through motion capture and plotted in real time in Fig.~\ref{fig:wind_resistance}. The extended wings configuration has a maximum positional deviation of 0.820 m. The retracted wings configuration shows a maximum displacement of 0.081 m. The significant reduction in positional offset strongly shows the effectiveness of the wing retraction system in enhancing the capability to reject wind disturbances.

\subsection{CHD Efficiency Verification}
\label{subsec:HDR_eff_ver}

The thrust and power of the propeller system are highly correlated with the propeller advance ratio $J$ and rotation speed $n$~\cite{brandt2011}, expressed by~\eqref{equ:porp_eff}
\begin{equation}
\label{equ:porp_eff}
\left\{ 
\begin{aligned}
    J &=\frac{V}{nD}, \\
    T &=C_T(J,n)\cdot\rho n^2 D^4, \\
    P &=C_P(J,n)\cdot\rho n^3 D^5 ,\\
    \end{aligned}
\right.
\end{equation}
where $T$ and $P$ are the thrust and the power of the propeller, respectively. $V$ is the speed of airflow, $D$ is the diameter of the propeller, $\rho$ is the air density. $C_T$ and $C_P$ are the thrust coefficient and power coefficient of the propeller, which are strongly related to the $J$ and $n$. 

To evaluate propulsion efficiency across heterogeneous and homogeneous configurations, three comparative simulation scenarios were designed: 

1) Heterogeneous propellers configuration (HPC): 5006 motor with 16-inch propeller and F90Pro motor with 7-inch propeller (the same configuration as the DART).

2) Homogeneous large-propeller configuration (HLC): dual 5006 motors with 16-inch propellers.

3) Homogeneous small-propeller configuration (HSC): dual F90Pro motors with 7-inch propellers. 

\begin{figure}[b]
    \centering
    \vspace{-12pt}
    \includegraphics[width=0.9\linewidth]{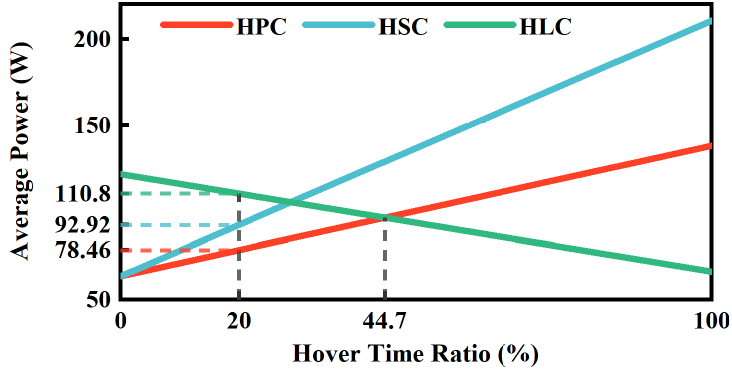}
    \vspace{-6pt}
    \caption{Average power consumption}
    \vspace{0pt}
    \label{fig:power_intersection}
\end{figure}
The DART (1.2 kg bare weight) requires 4 N thrust at 16 m/s cruise speed. Thrust settings were 12 N (multi-rotor) and 4 N (fixed-wing). Each propeller in HPC, HLC, and HSC provides ~6 N in multi-rotor mode for torque balance. In fixed-wing mode, both HPC and HSC use a single 7-inch propeller to generate the desired thrust. However, HLC is constrained to employ two propellers: a single MN5006 motor driving a 16-inch propeller delivers less than 4 N of thrust—insufficient for level flight—necessitating a dual-propeller configuration.
Using $C_T$, $C_P$ data of propellers from the UIUC Propeller Data Site~\cite{UIUC_PropDB}, we applied the~\eqref{equ:porp_eff} to calculate the 16-inch propeller's power consumption as 33 W in multirotor mode and 61 W in fixed-wing mode; The 7-inch propeller has a power of 105.3 W in multi-rotor mode and 63.5 W in fixed-wing mode.

As shown in Fig.~\ref{fig:power_intersection}, for the higher efficiency of 16-inch propellers in multi-rotor mode and 7-inch propellers in fixed-wing mode, HLC's average power decreases with increasing hover time ratio (the proportion of hovering time in total flight time), while HSC exhibits the opposite trend. The DART with HPC configuration demonstrates intermediate average power between HSC and HLC. In multi-rotor mode, HPC achieves $34\%$ lower average power than HSC. During fixed-wing mode, HPC demonstrates $48\%$ lower power consumption versus HLC. Notably, when the hover time ratio is below $44.7\%$, HPC consistently achieves lower power than both configurations. Specifically at $20\%$ hover ratio, a typical VTOL mission profile, HPC reduces average power by $15.6\%$ versus HSC and $29.2\%$ versus HLC, strongly validating the efficiency of the CHD configuration.

\subsection{Torque Sensor-Based Vibration Analysis of SPLM}
\label{subsec:viber_ver}
\begin{figure}[h]
    \centering
    \vspace{-6pt} 
    \includegraphics[width=0.9\linewidth]{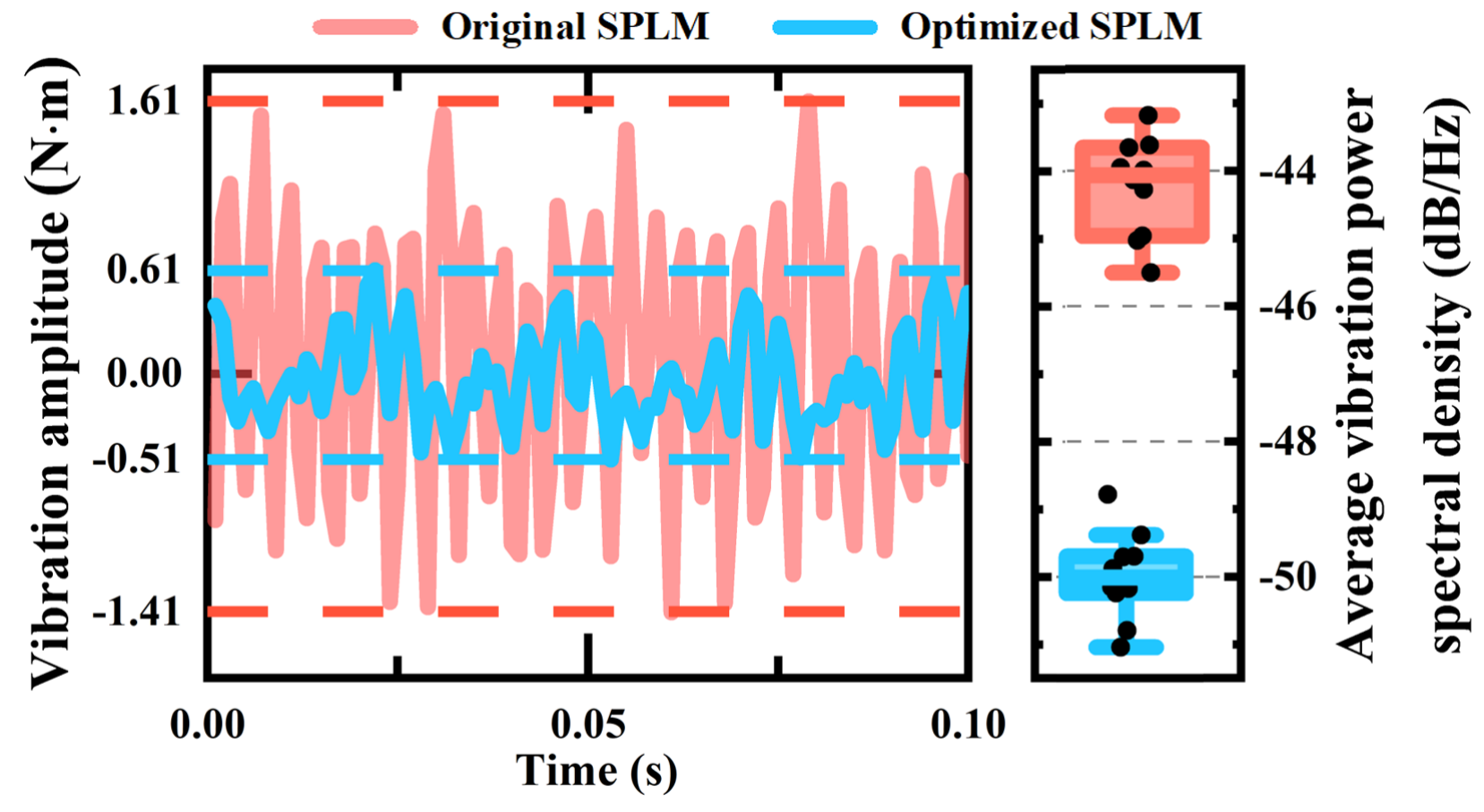}
    \vspace{-10pt} 
    \caption{Torque responses of SPLM configurations with and without flapping hinges}
    \label{fig:vibration_amp}
\end{figure}

\vspace{-3pt}
This section aims to verify the vibration metrics of the optimized SPLM configuration via a series of ablation experiments. In this experimental setup, the rotor was mounted on a torque sensor installed on a test bench. The $T_{d,1}$  was set to the DART hover throttle, specifically a fixed value of 900, with acceleration and deceleration maneuvers of fixed $||\mathbf{M}_d||$ (200) and phase (0). Torque data were acquired via the torque sensor at a sampling rate of 1000 Hz. Considering the slight discrepancies in the average torque magnitudes generated SPLMs, fixed window mean subtraction was applied to each dataset to analyze the uncertainty in torque response that leads to vibration. The processed resultant torque responses of two SPLMs over a specific sampling duration (0.1 s) were plotted in Fig.~\ref{fig:vibration_amp}. The torque sampling data of the optimized SPLM with flapping hinges exhibited a 62.9$\%$ reduction compared to the original SPLM without flapping hinges. To analyze the vibration characteristics in the frequency domain, ten identical experiments were conducted for two SPLM, and the data were subjected to PSD analysis. For each experiment, the PSD data were averaged over frequency to obtain the average vibration power spectral density. The results, summarized on the right of Fig.~\ref{fig:vibration_amp}, demonstrate that the average vibration PSD of the original SPLM is -44 dB/Hz, while the addition of flapping hinges reduces the average vibration PSD to -50 dB/Hz, representing a reduction of 6 dB/Hz. Experiments effectively validate the mitigation of SPLM vibrations with flapping hinges.

\subsection{Transition and Fixed-wing Mode Flight}
\label{subsec:trans}
\begin{figure}[h]
    \centering
    \vspace{3pt}
    \includegraphics[width=1\linewidth]{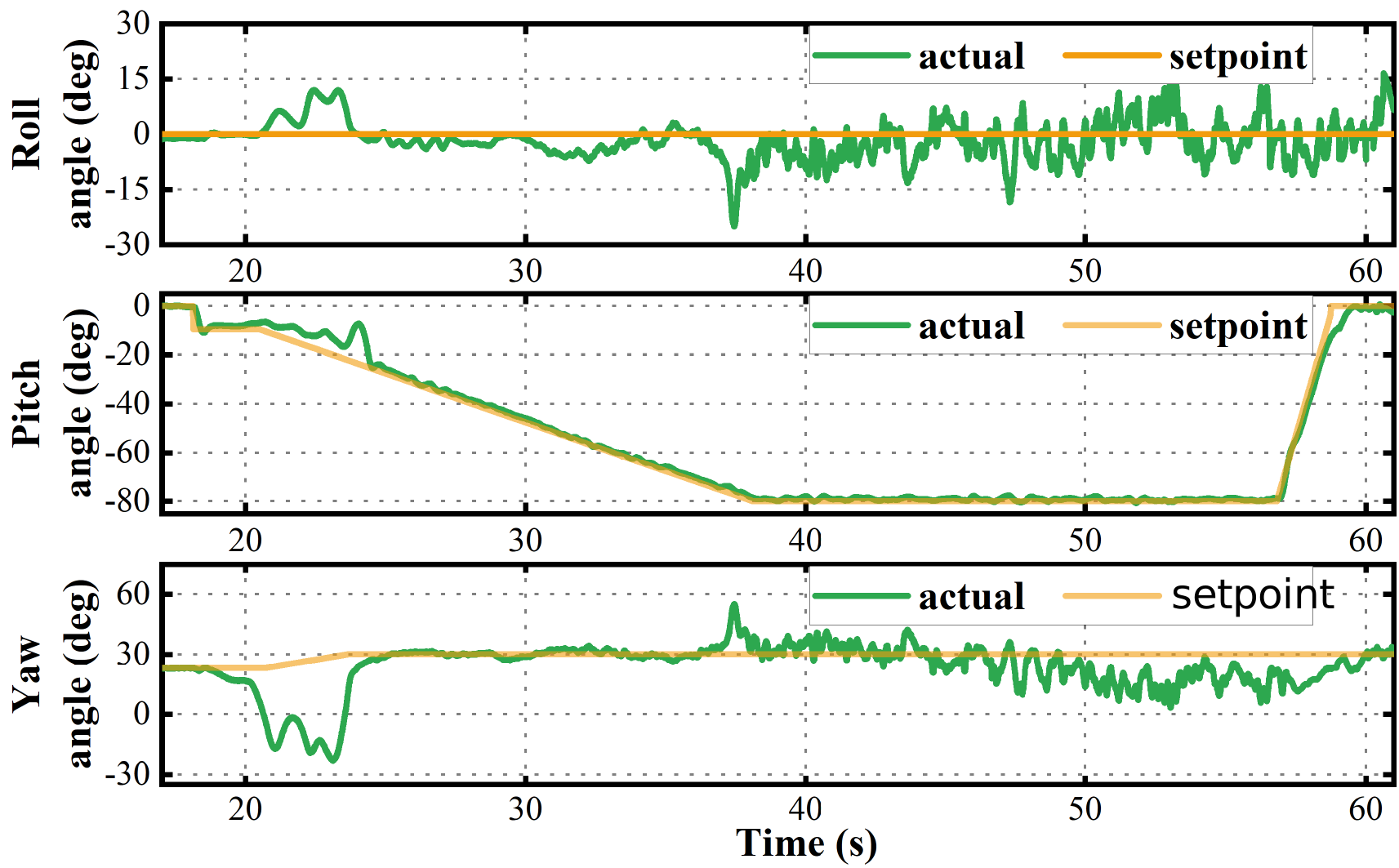}
     \vspace{-16pt}
    \caption{Tracking performance of roll, pitch, and yaw in outdoor flight test}
    \label{fig:trans_data}
    \vspace{0pt}
\end{figure}

We conducted transition and fixed-wing mode flight tests on the DART in an outdoor environment to validate its attitude tracking performance. The results are presented in Fig.~\ref{fig:trans_data}. At the beginning of the forward transition, the pitch setpoint is stepped to $-10^\circ$ and held for 2 seconds. It is then ramped linearly to $-80^\circ$ over the next 20 seconds, positioning the aircraft at its cruise AoA. Then, the vehicle remains in fixed-wing flight for 20 seconds, after which the pitch is ramped back to $0^\circ$ within 2 seconds to conclude the backward transition.
Under high-altitude complex and varying wind fields, roll and yaw angles exhibited minor oscillations around their setpoints. The pitch angle was tracked with a rapid response. In fixed-wing mode, the final ground speed reached 15.6 m/s, which matched the designed cruise speed, confirming a successful mode transition.

\section{CONCLUSION AND FUTURE WORK}
\label{sec:conclusion}

This paper presents the design and development of DART, a coaxial dual-rotor reconfigurable tailsitter UAV with an optimized swashplateless mechanism (SPLM). Its reconfigurable wings reduce frontal area by 66.2$\%$, the vehicle's susceptibility to wind disturbances was significantly reduced. The coaxial heterogeneous dual-rotor configuration lowers average power consumption by up to 29.2$\%$ in typical missions. The optimized SPLM with flapping hinges reduces vibration amplitude in the torque sampling data by 62.9$\%$ (6 dB/Hz in PSD). Flight tests validate transitions and fixed-wing cruise at 15.6 m/s.
Future work will refine SPLM dynamic modeling to enhance torque control accuracy, integrate autonomous flight for complex environments.
\bibliography{RA-L_25-3487_reference} 

@string{icra = {Proc. of the {IEEE} Intl. Conf. on Robot. and Autom.}}

@string{iros = {Proc. of the {IEEE/RSJ} Intl. Conf. on Intell. Robots and Syst.}}

@string{icuas = {Proc. of the Intl. Conf. on Unma. Air. Syst.}}

@article{bacchini2019electric,
  title={Electric VTOL configurations comparison},
  author={Bacchini, Alessandro and Cestino, Enrico},
  journal={Aerospace},
  volume={6},
  number={3},
  pages={26},
  year={2019},
  publisher={Multidisciplinary Digital Publishing Institute}
}

@inproceedings{chen2023swashplateless,
  title={Swashplateless-elevon actuation for a dual-rotor tail-sitter vtol uav},
  author={Chen, Nan and Kong, Fanze and Li, Haotian and Liu, Jiayuan and Ye, Ziwei and Xu, Wei and Zhu, Fangcheng and Lyu, Ximin and Zhang, Fu},
  booktitle={2023 IEEE/RSJ International Conference on Intelligent Robots and Systems (IROS)},
  pages={6970--6976},
  year={2023},
  organization={IEEE}
}

@article{cai2024development,
  title={Development of a tube-launched tail-sitter unmanned aerial vehicle},
  author={Cai, Jiaze and Denton, Hunter and Benedict, Moble and Kang, Hao},
  journal={International Journal of Micro Air Vehicles},
  volume={16},
  pages={17568293241254045},
  year={2024},
  publisher={SAGE Publications Sage UK: London, England}
}

@inproceedings{paulos2013underactuated,
  title={An underactuated propeller for attitude control in micro air vehicles},
  author={Paulos, James and Yim, Mark},
  booktitle={2013 IEEE/RSJ International Conference on Intelligent Robots and Systems},
  pages={1374--1379},
  year={2013},
  organization={IEEE}
}

@article{wang2015modeling,
  title={Modeling and control of an agile tail-sitter aircraft},
  author={Wang, Xinhua and Chen, Zengqiang and Yuan, Zhuzhi},
  journal={Journal of the Franklin Institute},
  volume={352},
  number={12},
  pages={5437--5472},
  year={2015},
  publisher={Elsevier}
}

@article{Chen2023ASS,
  title={A self-rotating, single-actuated UAV with extended sensor field of view for autonomous navigation},
  author={Nan Chen and Fanze Kong and Wei Xu and Yixi Cai and Haotian Li and Dongjiao He and Youming Qin and Fu Zhang},
  journal={Science Robotics},
  year={2023},
  volume={8},
  url={https://api.semanticscholar.org/CorpusID:257535202}
}

@ARTICLE{Gemini2,
  author={Qin, Youming and Chen, Nan and Cai, Yixi and Xu, Wei and Zhang, Fu},
  journal={IEEE/ASME Transactions on Mechatronics}, 
  title={Gemini II: Design, Modeling, and Control of a Compact Yet Efficient Servoless Bi-copter}, 
  year={2022},
  volume={27},
  number={6},
  pages={4304-4315},
  doi={10.1109/TMECH.2022.3153587}}

@inproceedings{paulos2018scalability,
  title={Scalability of cyclic control without blade pitch actuators},
  author={Paulos, James J and Yim, Mark},
  booktitle={2018 AIAA Atmospheric Flight Mechanics Conference},
  pages={0532},
  year={2018}
}

@inproceedings{paulos2018emulating,
  title={Emulating a fully actuated aerial vehicle using two actuators},
  author={Paulos, James and Caraher, Bennet and Yim, Mark},
  booktitle={2018 IEEE International Conference on Robotics and Automation (ICRA)},
  pages={7011--7016},
  year={2018},
  organization={IEEE}
}

@ARTICLE{sun2018wind,
  author={Sun, Jingxuan and Li, Boyang and Wen, Chih-Yung and Chen, Chih-Keng},
  journal={IEEE Transactions on Aerospace and Electronic Systems}, 
  title={Model-Aided Wind Estimation Method for a Tail-Sitter Aircraft}, 
  year={2020},
  volume={56},
  number={2},
  pages={1262-1278},
  doi={10.1109/TAES.2019.2929379}}

@inproceedings{simmons2022efficient,
  title={Efficient Variable-Pitch Propeller Aerodynamic Model Development for Vectored-Thrust eVTOL Aircraft},
  author={Simmons, Benjamin M},
  booktitle={AIAA Aviation 2022 Forum},
  pages={3817},
  year={2022}
}

@inproceedings{gao2024design,
  title={Design and verification of controller for a small coaxial dual-rotor aircraft},
  author={Gao, Botao and Wu, Zhilin and Wu, Mingjian and Li, Shuang},
  booktitle={Journal of Physics: Conference Series},
  volume={2764},
  number={1},
  pages={012014},
  year={2024},
  organization={IOP Publishing}
}

@article{wei2019research,
  title={Research on the control algorithm of coaxial rotor aircraft based on sliding mode and PID},
  author={Wei, Yiran and Chen, Han and Li, Kewei and Deng, Hongbin and Li, Dongfang},
  journal={Electronics},
  volume={8},
  number={12},
  pages={1428},
  year={2019},
  publisher={MDPI}
}

@inproceedings{robinson2013design,
  title={Design and system identification of a micro coaxial helicopter testbed},
  author={Robinson, DC and Doherty, Erin and Tsai, Steve and Chung, Hoam},
  booktitle={2013 IEEE/ASME International Conference on Advanced Intelligent Mechatronics},
  pages={1423--1428},
  year={2013},
  organization={IEEE}
}

@inproceedings{garcia2009modeling,
  title={Modeling and control of a vectored-thrust coaxial UAV},
  author={Garcia, Octavio and Sanchez, Anand and Wong, KC and Lozano, Rogelio},
  booktitle={2009 European Control Conference (ECC)},
  pages={695--700},
  year={2009},
  organization={IEEE}
}

@article{ang2015design,
  title={Design and implementation of a thrust-vectored unmanned tail-sitter with reconfigurable wings},
  author={Ang, Kevin ZY and Cui, Jin Q and Pang, Tao and Li, Kun and Wang, Kangli and Ke, Yijie and Chen, Ben M},
  journal={Unmanned Systems},
  volume={3},
  number={02},
  pages={143--162},
  year={2015},
  publisher={World Scientific}
}

@article{duan2024optimization,
  title={Optimization-Based Control for a Large-Scale Electrical Vertical Take-Off and Landing during an Aircraft’s Vertical Take-Off and Landing Phase with Variable-Pitch Propellers},
  author={Duan, Luyuhang and He, Yunhan and Fan, Li and Qiu, Wei and Wen, Guangwei and Xu, Yun},
  journal={Drones},
  volume={8},
  number={4},
  pages={121},
  year={2024},
  publisher={MDPI}
}

@inproceedings{gu2017development,
  title={Development and experimental verification of a hybrid vertical take-off and landing (VTOL) unmanned aerial vehicle (UAV)},
  author={Gu, Haowei and Lyu, Ximin and Li, Zexiang and Shen, Shaojie and Zhang, Fu},
  booktitle={2017 International Conference on Unmanned Aircraft Systems (ICUAS)},
  pages={160--169},
  year={2017},
  organization={IEEE}
}

@INPROCEEDINGS{Henderson2020optcontrol,
  author={Henderson, Travis and Papanikolopoulos, Nikolaos},
  booktitle={2020 IEEE International Conference on Robotics and Automation (ICRA)}, 
  title={Adaptive Control of Variable-Pitch Propellers: Pursuing Minimum-Effort Operation}, 
  year={2020},
  volume={},
  number={},
  pages={7470-7476},
  doi={10.1109/ICRA40945.2020.9197208}}

@inproceedings{paulos2015flight,
  title={Flight performance of a swashplateless micro air vehicle},
  author={Paulos, James and Yim, Mark},
  booktitle={2015 IEEE International Conference on Robotics and Automation (ICRA)},
  pages={5284--5289},
  year={2015},
  organization={IEEE}
}

@inproceedings{brandt2011,
  author    = {Brandt, J. and Selig, M.},
  title     = {Propeller performance data at low reynolds numbers},
  booktitle = {49th AIAA Aerospace Sciences Meeting including the New Horizons Forum and Aerospace Exposition},
  year      = {2011},
  pages     = {1255},
  address   = {Orlando, Florida},  
  month     = {Jan. 4-7},         
  doi       = {10.2514/6.2011-1255} 
}

@dataset{UIUC_PropDB,
  author = {Brandt, Jonathan B. and Deters, Robert W. and Ananda, Girish K. and Dantsker, Oleg D. and Selig, Michael S.},
  title = {{UIUC Propeller Database}, Volumes 1-4},
  howpublished = {University of Illinois at Urbana-Champaign, Department of Aerospace Engineering},
  url = {https://m-selig.ae.illinois.edu/props/propDB.html},
  year = {2015},          
  version = {3.0},         
  note = {Comprehensive experimental dataset containing 172 propeller geometries with aerodynamic performance measurements}
}

@article{vourtsis2021robotic,
  title={Robotic elytra: Insect-inspired protective wings for resilient and multi-modal drones},
  author={Vourtsis, Charalampos and Stewart, William and Floreano, Dario},
  journal={IEEE Robotics and Automation Letters},
  volume={7},
  number={1},
  pages={223--230},
  year={2021},
  publisher={IEEE}
}

@article{lyu2018disturbance,
  title={Disturbance Observer Based Hovering Control of Quadrotor Tail-Sitter VTOL UAVs Using $H_\infty$ Synthesis},
  author={Lyu, Ximin and Zhou, Jinni and Gu, Haowei and Li, Zexiang and Shen, Shaojie and Zhang, Fu},
  journal={IEEE Robotics and Automation Letters},
  volume={3},
  number={4},
  pages={2910--2917},
  year={2018},
  publisher={IEEE}
}
\end{document}